\pdfoutput=1

\documentclass{article}

\PassOptionsToPackage{numbers,sort&compress}{natbib}

 \usepackage[main, final]{neurips_2026}

\usepackage[utf8]{inputenc} 
\usepackage[T1]{fontenc}    
\usepackage{hyperref}       
\usepackage{url}            
\usepackage{booktabs}       
\usepackage{amsfonts}       
\usepackage{nicefrac}       
\usepackage{microtype}      
\usepackage{xcolor}         
\usepackage[capitalise]{cleveref}
\usepackage{enumitem}

\usepackage{adjustbox}
\usepackage{booktabs}
\usepackage{multirow}
\usepackage{graphicx}
\usepackage{wrapfig}
\usepackage{subcaption}

\title{Lost in the Slots: Revisiting Object-Centric Representations in the era of Foundation Models}

\author{%
  Priyam Dey\thanks{Equal Contribution. Correspondence at: {\tt\small priyamdey@iisc.ac.in, adityasahdev.97@gmail.com}.\\\hspace*{1.2em}$\dagger$ Work done as a Pre-doctoral researcher at Vision and AI Lab. Current Affiliation: Sarvam AI.\\\hspace*{1.2em}$^{\diamond}$ Work done as a Research Intern at Vision and AI Lab. Current Affiliation: University of California, San Diego.}
  \And
  \textbf{Aditya Sahdev$^{*\dagger}$}
  \And
  \textbf{Omkar Mahesh Kashyap$^{\diamond}$}
  \AND
  \textbf{R. Venkatesh Babu} \\[6pt]
  Vision and AI Lab, Indian Institute of Science, Bangalore \\[4pt]
  Project Page: {\small\url{https://val-iisc.github.io/lost-in-the-slots/}}
}

\begin{document}

\maketitle

\begin{abstract}
Object-centric representation learning (OCL) has long been proposed as a principled solution to the binding problem, promising improved compositional reasoning and robustness. Yet its effectiveness in the era of foundation models remains unclear. We systematically evaluate whether explicit object-centric (OC) representations improve binding by pairing slot-based representations with pretrained foundation LLMs and evaluating them on diverse open-ended VQA, grounding, and hallucination benchmarks. Contrary to common assumptions, we find that slot-based OC representations consistently underperform dense patch-based features across all settings. To understand these failures, we analyze slot representations through the lens of the core components of binding -- scene segregation and object representation -- and identify three key limitations: (i) \textit{inferior scene segregation} compared to the implicit grouping already present in pretrained visual backbones, (ii) \textit{inherent information loss} introduced during slot encoding that degrades downstream reasoning, and (iii) \textit{weak attribute encoding} of fine-grained object properties such as category, color, and spatial position. Extending our analysis beyond slot representations, we find that although dense visual representations exhibit stronger binding, they are themselves imperfect, with additional binding failures arising in the LLM. Overall, our findings challenge the practical utility of existing OCL methods in foundation models and suggest that future approaches should preserve rich visual information while introducing object-centric mechanisms that enable more reliable multimodal binding.
\end{abstract}    

\section{Introduction}


Humans perceive and interact with the world in terms of structured objects rather than disjoint features \citep{peters2021capturing,scholl2001objects}. A key mechanism underlying this object-centric abstraction is \textit{binding}: the ability to integrate distributed information into coherent entities and flexibly relate them during reasoning \citep{feldman2013neural,greff2020binding,von_der_malsburg1994correlation}. Humans possess highly effective binding mechanisms, yet they still struggle under challenging conditions. For example, in visual conjunction search, human performance deteriorates as the number of objects and shared features increases \citep{treisman1982illusory,campbell2024understanding}. Modern AI models exhibit related, and often more pervasive, binding failures, including errors in spatial reasoning and object and attribute hallucinations \citep{campbell2024understanding,assouel2025objectcentric}.

Object-centric representation learning (OCL) has emerged as a principled approach to this problem, wherein it explicitly decompose a scene into a set of object representations, commonly referred to as slots, with the goal of separating objects while retaining their attributes \citep{slot_attention_locatello_neurips2020,ding2021attention,seitzer2023bridging,didolkar2025on}. Such representations have demonstrated benefits for compositional generalization, robotics, video modeling, visual question answering, and generative modeling \citep{singh2022illiterate,pmlr-v202-driess23a,elsayed2022savi,oc_in_vqa_mamaghan_iclr2025,wu2023slotdiffusion}. More recently, object-centric representations have also been proposed as a potential architectural pathway toward addressing binding failures in vision-language models \citep{campbell2024understanding,visualsymbolic_in_vlm_assouel_iclr2026}. However, despite these successes, it remains unclear whether current slot-based object-centric (OC) representations actually provide a better substrate for binding and reasoning than the dense visual representations learned by modern foundation models.

\begin{figure}
    \centering
    \includegraphics[width=0.95\textwidth]{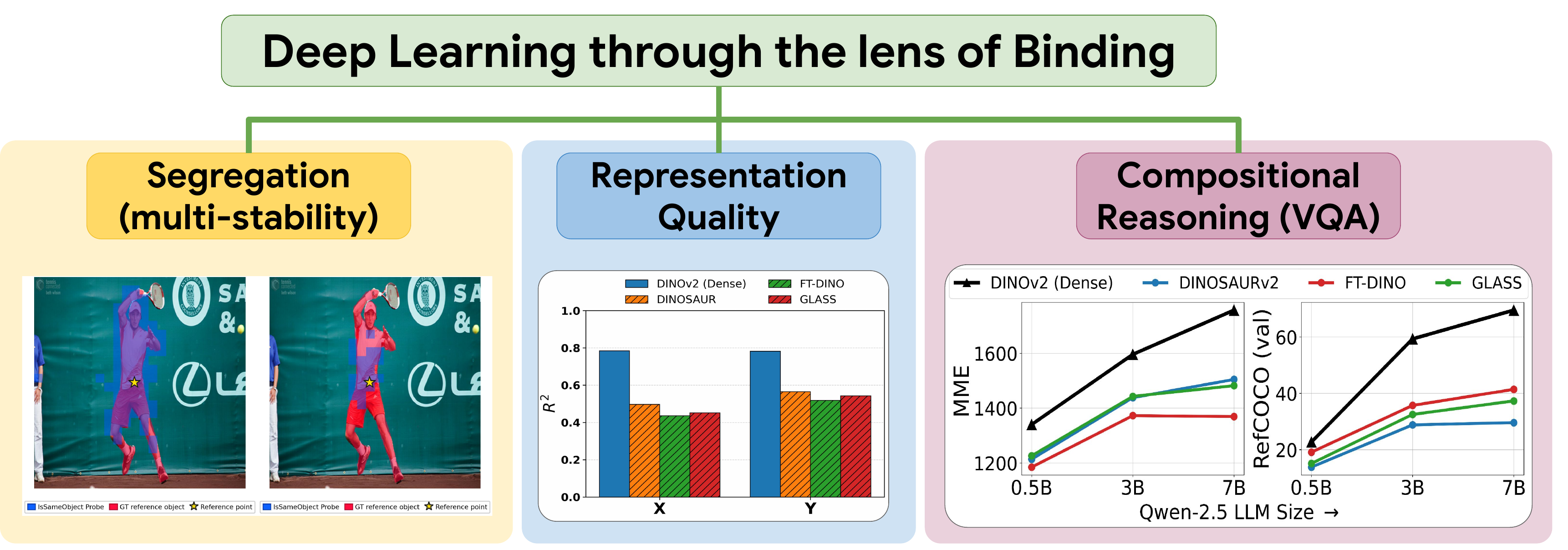}
    \caption{We analyze object-centric (OC) representations against deep visual features (e.g., of DINOv2 \citep{oquab2024dinov}) using the \textit{core functional blocks} in binding \citep{greff2020binding}: \textbf{(a) Segregation.} Slot-based OC methods provide a single-discrete decomposition, instead, dense patch representations of models like DINOv2 encode multi-stable grouping affinities (\cref{subsec:oc_seg_implicit_seg}). \textbf{(b) Representation.} Using property prediction task, we show that OC representations incur loss in information w.r.t dense patch representations, resulting in poor decodability of relevant object properties (\cref{subsec:repr_quality_oc}) \textbf{(c) Compositional Reasoning.}  We use VLMs as an interface to evaluate both OC and dense patch representations for compositional reasoning, finding that dense representations consistently outperform OC methods across model scales on various benchmarks (\cref{sec:main_eval}).}
    \label{fig:teaser}
\end{figure}

In this work, we investigate this question systematically through the three functional components of binding identified by \citet{greff2020binding}: \textit{scene segregation}, \textit{object representation}, and \textit{compositional reasoning} (\cref{fig:teaser}). Using them as anchor points, we design systematic and controlled experiments to analyze OC representations in detail and assess \textit{whether}, and \textit{to what extent}, these properties emerge in models with an explicit object-centric bias. Rather than evaluating OC models solely on unsupervised object discovery as is prevalent in the OC literature, we assess their representations within a realistic foundation-model setting. Following Cambrian-1 \citep{tong2024cambrian}, we leverage off-the-shelf pretrained foundation LLMs as a primary interface for evaluating the quality of OC representations. In doing so, we address the key methodological limitations of \cite{mamaghan2025exploring}: we pretrain OC models on a standard dataset with \textit{broad coverage} (e.g. COCO \cite{mscoco_eccv14}), integrate them with \textit{pretrained foundation LLMs}, and perform visual instruction tuning. The resulting VLMs are then evaluated on a diverse suite of VQA benchmarks using \textit{standard protocols}, namely, free-form answer generation with correctness assessment, thereby avoiding the restrictive fixed-set classification formulation of prior work. Under this setup, our analysis reveals a contrasting picture which we discuss next. Concurrent to our work, \cite{singh2026evaluatingobjectcentricmodelsobject} also evaluate OC models in a similar VLM setting, though with a different focus, scope and conclusion; we discuss both works in detail in \cref{sec:related_works} and \cref{sec:related_works_supp}
 of the supplementary.

Our analysis reveals that current slot-based object-centric representations consistently underperform dense visual representations across varied slot counts on a broad range of downstream tasks, including visual reasoning, grounding, and hallucination evaluation. Next, this gap reflects limitations in both scene segregation and object representation: existing slot representations do not reliably improve object separation over the implicit segregation \textit{already present} in pretrained dense features, while the slot-encoding process can \textit{discard} information that is useful for downstream reasoning. Third, object-centric representations are not the only source of binding failures. Even with strong dense visual features, foundation VLMs remain susceptible to interference from \textit{task-irrelevant} visual information. Targeted interventions show that these failures arise \textit{primarily within} the language model, suggesting that robust multimodal binding requires co-design across the visual and language components rather than improvements to the visual encoder \textit{alone}.

Together, our findings suggest that the limitations of current OCL systems are not simply a consequence of insufficient object-centric bias, rather, existing slot-based mechanisms impose a costly information bottleneck without reliably delivering the segregation and representation quality required for downstream reasoning. At the same time, the binding failures observed in dense-feature VLMs indicate that future object-centric approaches should not only focus on representing objects but also on \textit{object-centric mechanisms} that enforce selective binding through explicit bottlenecks while preserving task-relevant visual information. We summarize our contributions below:
\begin{itemize}[noitemsep, topsep=0pt, leftmargin=10pt]
    
    \item We show that current slot-based object-centric representations, when analyzed systematically through the lens of binding, consistently underperform standard dense patch-based representations, challenging the common assumption that explicit object-centric factorization inherently improves multimodal reasoning and binding.
    
    \item We provide a detailed analysis of slot representations through the lens of binding, identifying three core limitations: inferior scene segregation, information loss during slot encoding, and weak attribute preservation for fine-grained object properties.
    
    \item We demonstrate that while dense representations exhibit stronger emergent binding behavior, both dense and slot-based approaches remain imperfect, motivating future multimodal architectures and mechanisms that better balance structural reasoning with information preservation.
\end{itemize}
\section{Related Works}
\label{sec:related_works}

\noindent \textbf{Binding in Deep Learning.}
The binding problem -- associating attributes with the correct entities -- remains a fundamental challenge in deep learning, particularly in modern VLMs~\citep{vlm_limits_campbell_neurips2024, visual_structures_help_vlm_izadi_neurips2025}. Recent approaches mitigate binding failures through object grounding~\citep{obj_grounding_vlms_geigle-emnlp2024}, object-centric structure~\citep{oc_in_vqa_mamaghan_iclr2025}, or explicit structural and symbolic mechanisms~\citep{ctrl_o_Didolkar_cvpr2025, ocbinding_in_clip_assouel_neurips2025, visualsymbolic_in_vlm_assouel_iclr2026}. While these works suggest structured representations are beneficial, they stop short of establishing whether existing object-centric formulations are sufficient to address binding failures in practice.

\noindent \textbf{Object-Centric Representation Learning (OCL).}
Object-centric learning aims to decompose scenes into structured object-level representations, with approaches ranging from early variational and attention-based methods~\citep{attend_infer_repeat_eslami_neurips19, monet_burgess19, iodine_greff19a_icml19, space_Lin_iclr2020, genesis_Engelcke_iclr2020} to Slot Attention~\citep{slot_attention_locatello_neurips2020} and its extensions~\citep{slate_singh_iclr2022, scalor_Jiang_iclr2020}. DINOSAUR~\citep{dinosaur_seitzer_iclr2023} bridges the gap to real-world images by grounding slot representations in pretrained DINO features, with follow-up works introducing additional inductive biases~\citep{cae_lowe_neurips2023, spot_Kakogeorgiou_cvpr2024}. Several works have attempted to characterize the role of object-centric representations through robustness, invariance, and compositional generalization~\citep{pmlr-dittadi_icml2022, kapl2026are}.

\noindent \textbf{OCL in Vision-Language Models.}
Visual question answering -- reasoning about image content in response to natural language queries -- represents one of the most prominent and rapidly growing applications of vision-language models. Recent efforts have explored incorporating object-centric representations into VLMs, including emergent object-like groupings from language supervision~\citep{groupvit_Xu_cvpr2022}, object-aware reasoning in large multimodal systems~\citep{groundhog_Zhang_cvpr2024, omg_llava_neurips2024}, and slot-based representations for structured generation~\citep{slotdiffusion_neurips2023, oc_slot_diffusion_jiang_neurips2023}. Given its centrality as a testbed for visual reasoning, we use the VLM setting as a substrate to investigate whether OC representations succeed or fail at binding.
Most closely related, \citet{oc_in_vqa_mamaghan_iclr2025} also evaluate OC representations in a VQA setting, but train small language models from scratch on the same dataset used for evaluation, precluding reliable conclusions about generalization, and formulate VQA as closed-set classification rather than open-ended QA. \citet{singh2026evaluatingobjectcentricmodelsobject}, on the other hand, evaluate OC representations in a foundation VLM setting and study how object-discovery metrics relate to downstream performance; while we share a similar evaluation protocol, our work differs substantially in scope, focus, and conclusions -- we elaborate on all these differences, and provide a more detailed discussion of related work more broadly, in \cref{sec:related_works_supp} of the supplementary.

\begin{table*}
\centering
\caption{\textbf{Performance comparison of OCL methods with their dense counterparts} on various VQA benchmarks across different vision encoders and LLM backbones. RefCOCO results are presented as (val/+/g). Dense features outperform OC methods across VQA tasks and backbones.}
\setlength{\tabcolsep}{4pt} 
\label{tab:main_eval}

\resizebox{\linewidth}{!}{%
\begin{tabular}{l ccccc ccccc}
\toprule
\multirow{2}{*}{\textbf{Encoder}} 
& \multicolumn{5}{c}{\textbf{Phi-4 3.8B}} 
& \multicolumn{5}{c}{\textbf{Vicuna-7B}} \\
\cmidrule{2-11}

& \textbf{MME} 
& \textbf{POPE} 
& \textbf{CVBench} 
& \textbf{GQA} 
& \textbf{RefCOCO} 

& \textbf{MME} 
& \textbf{POPE} 
& \textbf{CVBench} 
& \textbf{GQA} 
& \textbf{RefCOCO} \\
\midrule

\multicolumn{11}{c}{\textbf{Backbone: DINOv1}} \\
\midrule

Dense
& \textbf{1422.9} & \textbf{80.2} & \textbf{58.4} & \textbf{54.4} & \textbf{46.2/37.2/42.4} 
& \textbf{1477.6} & \textbf{81.3} & 54.4 & \textbf{57.9} & \textbf{53.3/43.2/47.7} \\

SPOT
& 1377.2 & 79.9 & 54.4 & 48.8 & 25.5/17.0/21.9 
& 1395.2 & 78.7 & \textbf{55.2} & 54.3 & 23.2/16.4/22.1 \\

\midrule

\multicolumn{11}{c}{\textbf{Backbone: DINOv2}} \\
\midrule

Dense
& \textbf{1548.8} & \textbf{86.8} & \textbf{62.9} & \textbf{61.9} & \textbf{65.5/59.8/61.6} 
& \textbf{1660.7} & \textbf{86.8} & \textbf{59.2} & \textbf{63.1} & \textbf{67.4/58.7/63.3} \\

DINOSAURv2
& 1421.3 & 81.8 & 57.8 & 53.7 & 31.0/21.0/37.9 
& 1509.6 & 82.6 & 56.5 & 56.8 & 35.5/29.8/34.0 \\

FT-DINO
& 1379.5 & 82.5 & 55.5 & 54.3 & 43.9/34.8/39.6 
& 1436.5 & 82.3 & 54.7 & 56.5 & 45.1/40.7/43.4 \\

GLASS
& 1478.3 & 81.6 & 57.3 & 53.4 & 39.3/27.7/32.1 
& 1653.7 & 82.4 & 57.4 & 56.8 & 39.7/33.3/37.5 \\

\bottomrule
\end{tabular}
}
\end{table*}

\section{How Good are OC Representations? Evidence from Foundation VLMs}
\label{sec:main_eval}
As discussed earlier, OC representations are currently primarily evaluated using unsupervised object discovery benchmarks, where representations are assessed based on how well they reconstruct the segmentation masks of the objects they spatially correspond to. Such evaluation protocols are inherently limiting and does not showcase the actual efficacy of the OC representations. Therefore, we move beyond this setup and leverage VLMs as an interface to evaluate their utility. This approach enables us to probe a broad range of properties of the underlying visual representations through diverse VQA benchmarks available in the literature, including the ones pertinent to binding.


\noindent \textbf{Setup.} We adopt the standard LLaVA visual instruction-tuning protocol \citep{liu2023visual} to integrate visual representations (object-centric or otherwise) with foundation LLMs. This involves a two-stage training process: in Stage I, a projection MLP is trained to map visual features to the LLM embedding space while keeping the vision encoder and LLM frozen; in Stage II, both the projector and the LLM are jointly fine-tuned using visual instruction data to enable effective multimodal reasoning. The resulting VLM is then evaluated on downstream benchmarks relevant to binding. Detailed description of the training datasets are provided in \cref{subsec:vlm_instruct_tuning_details_supp}. For our analysis, we consider four recent OC methods, including \textit{state-of-the-art} approaches: DINOSAUR \citep{seitzer2023bridging}, FT-DINO \citep{didolkar2025on}, SPOT \citep{spot_Kakogeorgiou_cvpr2024}, and GLASS \citep{glass_Singh_2025_CVPR}. We compare their VQA performance against corresponding dense representations across two vision backbones (DINOv1 \citep{Caron_2021_ICCV} / v2 \citep{oquab2024dinov}), three foundation LLMs (Phi-$4$ \citep{abdin2024phi4technicalreport}, Vicuna \citep{vicuna2023}, and Qwen-$2.5$ \citep{qwen2025qwen25technicalreport}), and multiple model scales ($0.5$B, $4$B, and $7$B)\footnote{Not all model families provide all size variants; we use the configurations available publicly.}.

\begin{table*}
\centering
\caption{\textbf{Performance comparison of OCL methods with their dense counterparts} across different vision encoders with Qwen2.5 LLM backbones (0.5B, 3B, and 7B) on various VQA benchmarks. RefCOCO results are presented as (val/+/g), and CVB stands for CV-Bench \cite{tong2024cambrian}. Due to lack of space, we report GQA \cite{Hudson_2019_CVPR} results in the supplementary. Dense features consistently outperform OCL-based methods across scales and backbones, with OCL methods witnessing poor scaling.}
\setlength{\tabcolsep}{1pt}
\label{tab:qwen_eval}
\resizebox{\linewidth}{!}{%
\begin{tabular}{l cccc cccc cccc}
\toprule
\multirow{2}{*}{\textbf{Encoder}} 
& \multicolumn{4}{c}{\textbf{Qwen2.5-0.5B}} 
& \multicolumn{4}{c}{\textbf{Qwen2.5-3B}}
& \multicolumn{4}{c}{\textbf{Qwen2.5-7B}} \\
\cmidrule{2-13}
& \textbf{MME} & \textbf{POPE} & \textbf{CVB} & \textbf{RefCOCO} 
& \textbf{MME} & \textbf{POPE} & \textbf{CVB} & \textbf{RefCOCO}
& \textbf{MME} & \textbf{POPE} & \textbf{CVB} & \textbf{RefCOCO} \\
\midrule

\multicolumn{13}{c}{\textbf{Backbone: DINOv1}} \\
\midrule
DINOv1
& \textbf{1170.1} & \textbf{80.6} & \textbf{41.7} & \textbf{22.7/18.7/21.7} 
& \textbf{1384.4} & \textbf{81.7} & \textbf{51.5} & \textbf{38.3/30.0/33.5}
& \textbf{1489.3} & \textbf{81.8} & \textbf{61.2} & \textbf{46.5/39.7/45.5} \\

SPOT  
& 1157.3 & 78.4 & 40.7 & 11.8/5.5/9.4 
& 1358.6 & 79.2 & 48.8 & 24.8/15.1/20.5
& 1442.0 & 78.8 & 58.8 & 27.1/19.2/24.0 \\

\midrule
\multicolumn{13}{c}{\textbf{Backbone: DINOv2}} \\
\midrule
DINOv2
& \textbf{1340.1} & \textbf{86.5} & 44.3 & \textbf{22.7/18.7/21.7}
& \textbf{1596.2} & \textbf{87.0} & \textbf{56.7} & \textbf{59.3/51.1/56.2}
& \textbf{1757.5} & \textbf{87.9} & \textbf{63.8} & \textbf{69.5/62.3/62.2} \\

DINOSAURv2 
& 1213.5 & 81.5 & 45.0 & 13.8/8.8/13.1 
& 1438.3 & 82.6 & 53.0 & 28.8/21.2/22.9
& 1504.7 & 82.6 & 60.4 & 29.6/24.2/29.0 \\

FT-DINO 
& 1184.6 & 81.8 & 41.7 & 19.1/13.8/16.3 
& 1372.6 & 82.9 & 53.5 & 35.7/28.3/33.2
& 1369.4 & 82.4 & 60.8 & 41.4/32.6/37.2 \\

GLASS
& 1226.2 & 79.6 & \textbf{46.7} & 15.1/10.3/12.9 
& 1442.9 & 82.0 & 54.3 & 32.5/24.2/26.9
& 1482.0 & 82.6 & 60.4 & 37.3/27.6/31.0 \\

\bottomrule
\end{tabular}
}
\end{table*}

\noindent \textbf{Results.} We report results in \cref{tab:main_eval} (Phi-$4$ and Vicuna) and \cref{tab:qwen_eval} (Qwen-$2.5$). A consistent pattern emerges across all settings: while upgrading the vision backbone (DINOv1 $\rightarrow$ DINOv2) yields substantial gains for dense patch representations, the same improvement is not observed for OC representations. For example, GLASS, one of the state-of-the-art OC methods built on DINOv2, performs only marginally better than, or on par with, DINOv1 on MME and POPE, while remaining significantly worse w.r.t its underlying DINOv2 backbone across all benchmarks. This gap is even more pronounced on grounding tasks: DINOv2 achieves $65.5$/$59.8$/$61.6$ on RefCOCO, whereas FT-DINO drops to $43.9$/$34.8$/$39.6$, which is worse than the DINOv1 dense features. This suggests that the OCL bottleneck introduces a lossy compression of spatial features critical for localization. A detailed breakdown of the results is provided in \cref{suppsec:additional_vqa_results} of the supplementary. 

A potential explanation for this performance degradation could be the limited number of slots (typically $7$\footnote{The choice of $7$ slots follows prior OC literature \citep{dinosaur_seitzer_iclr2023,didolkar2025on}, where it is a standard training hyperparameter and has been found to work well on datasets on which they're trained (such as COCO, which contain fewer than $10$ objects per image on average).}). To disentangle slot capacity from representation quality, we progressively increase the number of slots (e.g., to $64$ and $144$) for FT-DINO, which has been shown to generalize well to higher slot counts despite being trained with $7$ slots \citep{didolkar2025on}, which is sufficient to cover all objects in the scene relevant for the question along with the background, and report the results in \cref{tab:ftdino_eval_slot_count}. One can observe that, even with increased slot count, OC representations of FT-DINO continue to underperform their dense counterparts and also witnesses a saturation in performance, indicating that the gap is not merely due to slot capacity but is intrinsic to the slot encoding process. 

Further, scaling the LLM backbone exacerbates this gap rather than closing it: better LLMs consistently amplify the advantage of higher-quality dense features but provide limited benefit to OC (see \cref{fig:teaser}(c)). Taken together, these results indicate that OCL methods disrupt the synergy between vision encoders and LLMs: instead of benefiting from scaling, they saturate early, preserving coarse semantics while discarding the fine-grained structure needed for reasoning and localization.

\begin{figure}
    \centering
    \includegraphics[width=\textwidth]{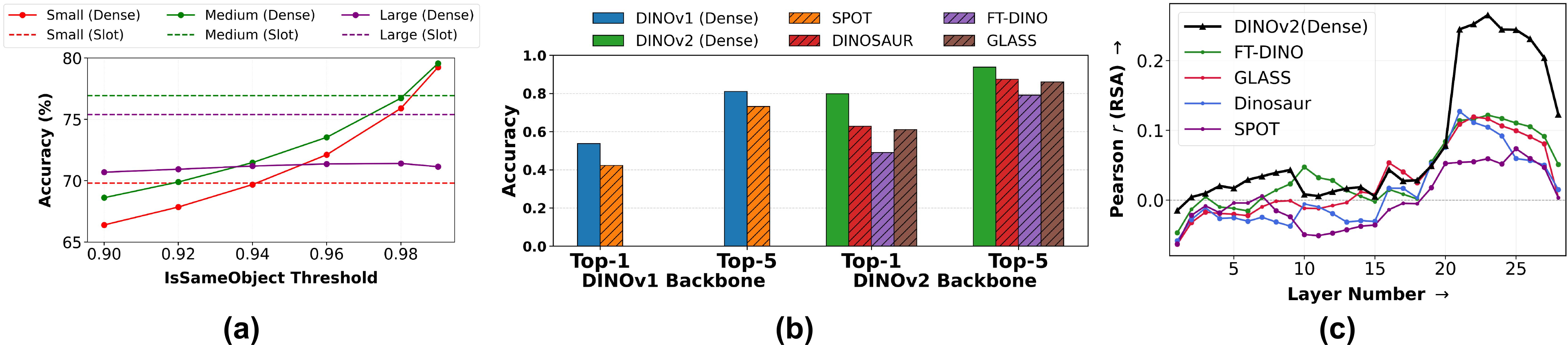}
    \caption{\textbf{OC Failure Analysis from the lens of Binding.} \textbf{\textit{(a)} Segmentation.} Dense patch representations achieve higher discriminability than all slot-based methods across object scales, with the gap persisting even against FT-Dino, the strongest OC baseline, beyond which other OC methods fall even further short.\textbf{\textit{(b)} Representation Quality.} We evaluate OC vs dense representations for property prediction task, observing that OC representations consistently underperform their dense counterparts in decoding relevant properties (category, position) of the underlying object. \textbf{\textit{(c)} RSA Analysis.} To understand impact of OC representations in a VLM, we perform RSA analysis \citep{visualsymbolic_in_vlm_assouel_iclr2026} of object position against hidden states across layers. Dense representations show a stronger  positional correlation, which is necessary for successful binding}
    \label{fig:sec4_combined}
\end{figure}

\section{Why Object-Centric Representations Fail?}
To understand why OCL methods consistently lag behind dense patch-based representations, we now investigate the fundamental, low-level properties of their underlying representations from the perspective of binding (\textit{segregation}, \textit{representation}), and report our findings. 

\subsection{OC Segmentation is Suboptimal to Implicit Segregation of Dense features}
\label{subsec:oc_seg_implicit_seg}

A central motivation behind slot-based OC representations is scene segregation: decomposing a visual scene into coherent object-level units through explicit object masks and slot representations. However, recent work \citep{li2026does} suggests that self-supervised vision transformers already encode rich object structure \textit{implicitly} within dense patch features without explicit grouping mechanisms. Therefore, we investigate whether this implicit segregation is more informative than the explicit segmentation produced by slot-based OC models, and what this implies for the utility of the OC inductive bias.


\textbf{Setup.} To quantify segregation quality in implicit dense representations, we employ the \textit{IsSameObject} probe of \citet{li2026does}, a lightweight quadratic classifier trained on ADE20K dataset \citep{Zhou_2017_CVPR} that predicts whether two patch-level features belong to the same object instance: Given two patch embeddings $\mathbf{x}_i$ and $\mathbf{x}_j$, the probe computes a similarity score $s=\sigma(\mathbf{x}_i^\top W\mathbf{x}_j+b)$, where $W$ and $b$ are learned parameters. A pair is labeled ``same'' object if both patches fall within the same ground-truth instance mask, and different object otherwise. For dense representations, the probe captures the implicit grouping structure encoded in the feature space, where patches from the same object are expected to exhibit higher similarity than patches from different objects. A pair is predicted as ``same'' when its pretrained DINOv2 feature similarity exceeds a fixed threshold. For OC representations, we derive an analogous grouping signal from the model's slot assignments: a pair is predicted as ``same'' when both patches are assigned to the same slot. We evaluate both approaches on $500$ COCO validation images \citep{mscoco_eccv14}, sampling $128$ patches distributed proportionally across its object instances, with each image containing at least $5$ annotated objects. We report accuracy for small, medium, and large objects following the standard COCO size definitions.


\textbf{Results and Discussion.} From \cref{fig:sec4_combined}(a), we observe that the \textit{IsSameObject} probe on dense patch representations achieves higher discriminability between same- and different-object patch pairs than slot-based grouping for all object sizes, with the only exception of FT-DINO on large objects. This advantage is particularly pronounced when the probe similarity threshold is adapted to object scale: higher thresholds capture finer objects, while lower thresholds capture larger regions. Such \textit{scale-adaptive} behavior highlights a key property of dense representations -- they encode a continuous spectrum of grouping affinities rather than committing to a fixed scene partition (for e.g., see \cref{fig:teaser}, multi-stability), which, as noted by \citet{greff2020binding}, is necessary for effective scene segregation.

In contrast, slot-based methods enforce a \textit{single discrete} decomposition through a fixed number of slots, limiting their ability to adapt to varying object scales at inference time. We note that even with an optimally chosen fixed threshold, dense representations remain competitive with or superior to slot-based segmentation, suggesting that the explicit grouping imposed by slot attention provides little advantage over the implicit object structure already present in pretrained dense representations. We corroborate this quantitatively across different OC methods and object sizes in \cref{tab:dense_vs_oc}, where we observe that dense representations consistently outperform OC methods on medium and small objects, with FT-DINO on large objects being the only exception.

\begin{table}
\centering
\caption{\textbf{Comparison of slot and patch-averaged representations across varying slot counts and methods.} Patch-average features are obtained by mean-pooling DINOv2 patch features within the segmentation mask of each slot from the respective method.  X and Y in property prediction denote the $R^2$ scores for the normalized horizontal and vertical coordinates, respectively. Patch-average \textit{outperforms} slot representations, indicating information loss in OC features relative to the dense counterparts, while increasing the number of slots \textit{does not yield} any consistent improvements.}
\label{tab:ftdino_eval_slot_count}
\vspace{0.1cm}
\setlength{\tabcolsep}{4pt}












\resizebox{0.95\linewidth}{!}{%
\begin{tabular}{ll|ccc|cccc}
\toprule
\multirow{2}{*}{\textbf{\# Slots}} & \multirow{2}{*}{\textbf{Representation}} & \multicolumn{3}{c|}{\textbf{VQA}} & \multicolumn{4}{c}{\textbf{Property Prediction}} \\
\cmidrule(lr){3-5} \cmidrule(lr){6-9}
 &  & \textbf{MME} & \textbf{POPE} & \textbf{CVBench} & \textbf{Top-1 (\%)} & \textbf{Top-5 (\%)} & \textbf{X ($R^2$)} & \textbf{Y ($R^2$)} \\
\midrule

\multicolumn{9}{c}{\textit{Method: DINOSAURv2}} \\
\midrule

\multirow{2}{*}{7}
& Slot & 1509.57 & \textbf{82.58} & \textbf{56.46} & \textbf{62.83} & 87.46 & \textbf{0.50} & \textbf{0.56} \\
& Patch-Average & \textbf{1536.76} & 82.40 & 56.15 & 62.80 & \textbf{87.70} & \textbf{0.50} & \textbf{0.56} \\

\midrule

\multicolumn{9}{c}{\textit{Method: FT-DINO}} \\
\midrule

\multirow{2}{*}{7}
& Slot & 1436.45 & 82.30 & 55.49 & 49.06 & 79.10 & 0.44 & \textbf{0.52} \\
& Patch-Average & \textbf{1614.33} & \textbf{82.96} & \textbf{58.52} & \textbf{61.20} & \textbf{86.10} & \textbf{0.50} & 0.51 \\

\midrule

\multirow{2}{*}{64}
& Slot & 1433.44 & 83.66 & 57.26 & 66.39 & 89.90 & 0.65 & 0.69 \\
& Patch-Average & \textbf{1584.30} & \textbf{84.29} & \textbf{58.20} & \textbf{74.60} & \textbf{91.00} & \textbf{0.71} & \textbf{0.71} \\

\midrule

\multirow{2}{*}{144}
& Slot & 1452.52 & 83.16 & 56.85 & 68.90 & 90.40 & 0.66 & 0.69 \\
& Patch-Average & \textbf{1600.23} & \textbf{84.06} & \textbf{58.66} & \textbf{77.70} & \textbf{93.79} & \textbf{0.72} & \textbf{0.73} \\

\bottomrule
\end{tabular}
}
\end{table}

\subsection{Understanding Representation Quality of OC methods}
\label{subsec:repr_quality_oc}
Given that poor segmentation quality of OC methods w.r.t the implicit segregation of the dense features, we now ask the same question for its representation quality: \textit{Are OC representations better than their dense counterparts on which they are built?} 

\textbf{Property Prediction.} We evaluate property prediction on COCO \citep{mscoco_eccv14}, following \citet{pmlr-dittadi_icml2022}. A linear predictor is trained to predict each object's category and normalized spatial coordinates from its representation. We report Top-1/Top-5 accuracy for category prediction and $R^2$ for coordinate prediction.
For OC representations, prediction is performed independently per slot, with Hungarian matching to ground-truth objects. For dense representations, we sample patches from the six largest ground-truth object masks and use the corresponding object category and patch-center coordinates as targets. Further details are provided in \cref{suppsec:property_prediction_task} of the supplementary.

The results are shown in \cref{tab:ftdino_eval_slot_count}. Across all evaluated OC methods, representations consistently underperform their dense counterparts on category and position prediction. To disentangle compression from slot encoding, we additionally evaluate a \textit{patch-average} baseline, where dense patch features corresponding to a slot mask are averaged to form a single object representation. We find that patch-averaged representations substantially outperform their slot-based counterparts, suggesting that the degradation arises not only from compression but also from the slot encoding process itself. 



\textbf{VQA evaluation of reconstructed OC features.} We further assess whether OC methods preserve information from the underlying dense representations by evaluating the dense features reconstructed by their slot decoders on downstream VQA benchmarks. Directly applying a VLM trained on the original dense features to these reconstructions results in near-random performance, indicating a substantial feature-space mismatch. We therefore instruction-tune the VLM on the reconstructed features using the same setup mentioned in \cref{sec:main_eval}, ensuring that each representation is evaluated within its appropriate input distribution. 

As shown in \cref{tab:reconstruction_eval}, reconstructed OC features generally underperform the corresponding dense representations across MME, POPE, and CVBench, with SPOT on MME being the only exception. In particular, the gap persists when moving from DINOv1 to DINOv2: although the underlying dense representations improve substantially, current OC methods fail to preserve or recover these gains through the slot encoding--decoding pipeline. These results further suggest that information lost during slot encoding cannot be fully recovered by the decoder.


\textbf{Representational Similarity Analysis.} To understand why OC representations underperform their dense counterparts, we examine how they are processed internally by the LLM, focusing on whether they induce different signatures of spatial binding within the LLM representations. We adapt the RSA \citep{kriegeskorte2008representational} framework of \citet{visualsymbolic_in_vlm_assouel_iclr2026}, who show that VLMs employ a two-stage symbol-like binding process: intermediate LLM layers first encode a content-independent spatial index of the target object, which is subsequently used in later layers to retrieve its features. Their RSA analysis measures the layer-wise correlation between the hidden state at the final token position and the target object's position or features. We adapt this analysis from synthetic scenes to natural images using the $2$D spatial reasoning subset of CV-Bench ($197$ images), and ask questions related to spatial relations, e.g., ``\textit{Select from the following choices. (A) left (B) right. Answer with the option's letter directly. Where is the person located with respect to the car?}'' For each image, we identify the queried object from the question (e.g., the person), detect it using OWLv2, and use the normalized center coordinates of its bounding box as its real-world position. Following \citet{visualsymbolic_in_vlm_assouel_iclr2026}, we compute, at each LLM layer, the Pearson correlation between the residual-stream representation at the question-mark token and the corresponding object position across the dataset, separately for dense and OC visual representations. This allows us to compare the spatial binding signatures induced by dense and OC representations and provides insight into how their differing representations are processed internally by the LLM.

\begin{table}

\begin{minipage}{0.54\linewidth}

\caption{\textbf{Performance comparison of reconstructed OC features} against the original dense representations across multiple VQA benchmarks. Reconstructed OC features consistently underperform the underlying dense representations, with SPOT on MME being the only exception. IT: instruction-tuned.}
\vspace{0.1cm}
\setlength{\tabcolsep}{2pt}
\label{tab:reconstruction_eval}

\resizebox{\linewidth}{!}{%
\begin{tabular}{lccc}
\toprule
\textbf{Encoder} & \textbf{MME} & \textbf{POPE} & \textbf{CVBench} \\
\midrule

DINOv1 & 1477.62 & \textbf{81.32} & \textbf{54.39} \\
SPOT Recon. (IT) & \textbf{1561.10} & 81.14 & 53.03 \\

\midrule

DINOv2 & \textbf{1660.68} & \textbf{86.79} & \textbf{59.17} \\
DINOSAURv2 Recon. (IT) & 1563.50 & 82.16 & 55.79 \\
FT-DINO Recon. (IT) & 1281.71 & 83.07 & 54.64 \\

\bottomrule
\end{tabular}}
\end{minipage}
\hfill
\begin{minipage}{0.44\linewidth}

\caption{\textbf{Comparison of dense and OC representations on the \textit{IsSameObject} probe across object sizes.} Dense representations consistently outperform slot-based OC methods across small, medium, and large objects, with FT-DINO on large objects being the only exception.}
\label{tab:dense_vs_oc}
\resizebox{1.0\linewidth}{!}{
\begin{tabular}{lccc}
\toprule
Method & Large & Medium & Small \\
\midrule
Dense    & 71.14          & \textbf{79.55} & \textbf{79.24} \\
\midrule
FT-DINO  & \textbf{75.39} & 76.93          & 69.79          \\
DINOSAUR & 68.26          & 76.74          & 74.65          \\
GLASS    & 68.96          & 67.16          & 62.06          \\
\bottomrule
\end{tabular}}
\end{minipage}
\end{table}

As shown in \cref{fig:sec4_combined}(c), the results reveal a striking gap. Dense DINOv2 representations exhibit a strong and sharply localised correlation peak ($r \approx 0.25$) in layers $21$--$24$, consistent with the emergence of a spatial index for the referenced object. All OC representations show a much weaker signal ($r < 0.13$), with correlations that emerge only partially and later in the network. This suggests that when OC-encoded features are used, the LLM struggles to form the content-independent positional indices that underpin successful binding. 

\subsection{What does a good OC method really need?}
\label{subsec:good_oc_method}
So far, we have identified two shortcomings of current OC methods: poor scene segregation and weak representations. We now turn to the converse question: \textit{what does a good OC method really need?} In particular, are both accurate scene segregation and high-quality representations necessary for strong downstream performance? Prior work \citep{mamaghan2025exploring,singh2026evaluatingobjectcentricmodelsobject} reports a weak correlation between standard unsupervised object-discovery metrics and downstream VLM performance. However, such correlations do not establish whether good scene segregation is itself necessary for strong downstream reasoning, since existing OC methods may simultaneously differ in representation quality. We therefore address this question through a controlled study that isolates the contribution of segmentation quality while holding representation quality fixed. We compare DINOSAUR \citep{seitzer2023bridging} and FT-DINO \citep{didolkar2025on}, which exhibit substantially different segmentation quality (FT-DINO $\gg$ DINOSAUR, AR/mBO reported in \cref{tab:supp_ar_mbo} of the supplementary), while fixing the representation mechanism. Specifically, we retain their predicted masks but replace their learned slot representations with patch-averaged frozen DINOv2 features (a mathematical description of this procedure is provided in \cref{suppsec:patch_average} of the supplementary). Crucially, this controlled setup varies segmentation quality while keeping the underlying visual representation fixed.

\textbf{Results.} As shown in \cref{tab:ftdino_eval_slot_count} (see results corresponding to $7$ slots), the patch-averaged representations derived from FT-DINO's higher-quality masks consistently outperform those derived from DINOSAUR's masks on VQA, providing evidence that, when representation quality is held fixed, better scene segregation \textit{does indeed improve} downstream reasoning. Conversely, under the same FT-DINO segmentation, patch-averaged DINOv2 features substantially outperform the learned slot representations, highlighting the independent importance of representation quality.

Together, these results suggest that segmentation and representation quality are complementary factors in object-centric VLMs, and in this particular scenario, we observe that FT-DINO achieves substantially better segmentation than DINOSAUR, yet its learned representations are weaker; replacing these representations with patch-averaged DINOv2 features recovers the benefit of its better segmentation. Thus, effective OC methodologies require both accurate scene segregation and preservation of the information contained in the underlying visual features for strong downstream reasoning capabilities.
\section{Binding in Foundation VLMs: Are Dense Representations Good Enough?}

\begin{table}

\begin{minipage}{0.45\linewidth}

\caption{
\small{\textbf{Binding analysis in Foundation VLMs: Task-Irrelevant Blackout (\cref{subsec:blackout}).} Blacking out task-irrelevant objects consistently improves performance, indicating that task-irrelevant visual content interferes with compositional reasoning.}}
\vspace{0.4cm}
\resizebox{1.0\linewidth}{!}{
\label{tab:binding_results}
\setlength{\tabcolsep}{4pt}
\centering
\begin{tabular}{@{} l c c c @{}}
\toprule
\textbf{Model} & \textbf{Condition} & \textbf{POPE} & \textbf{CV-Bench} \\
\midrule
\multirow{2}{*}{LLaVA-1.5}
  & Original          & 80.5 & 64.7 \\
  & +Blackout       & \textbf{81.7} & \textbf{65.8} \\
\addlinespace
\multirow{2}{*}{Qwen-2.5-VL}
  & Original          & 78.7 & 79.3 \\
  & +Blackout       & \textbf{81.9} & \textbf{80.2} \\
\bottomrule
\end{tabular}}
\end{minipage}
\hfill
\begin{minipage}{0.52\linewidth}
\centering
          \includegraphics[width=0.85\linewidth]{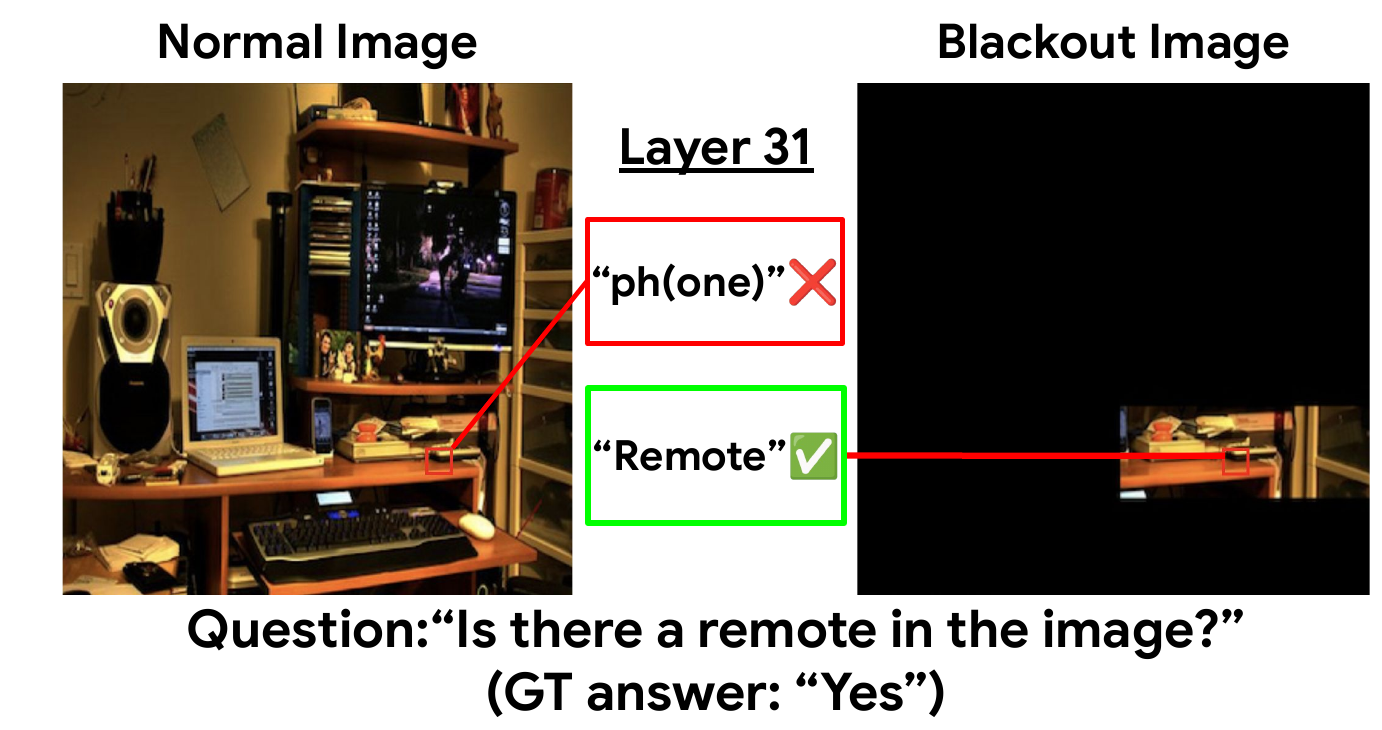}
        \captionof{figure}{\small{\textbf{Logit Lens analysis of Task-Irrelevant blackout.} As we omit out information irrelevant to the object of interest, the model answers correctly due to higher confidence on the right answer, as shown above by an example of ``remote'' in an indoor scene of cluttered objects.}}
        \label{fig:logit_lens}
        
\end{minipage}

\end{table}

Having established the shortcomings of slot-based OC representations, we now turn to a more fundamental question: \textit{Do the dense visual representations underlying OC models themselves support robust binding?} To investigate this, we directly probe binding failures in foundation VLMs built on dense patch representations through controlled interventions on the visual input, analyzing their effects on downstream performance and internal model behavior. In this framework, VLMs serve both as a diagnostic tool for evaluating the binding capacity of dense representations and as objects of study for understanding and mitigating the mechanisms underlying their binding failures.

\subsection{Do Task-Irrelevant Objects Interfere with Compositional Reasoning?}
\label{subsec:blackout}

Even when a visual encoder produces rich and expressive representations, a VLM must selectively bind the information relevant to the question at hand while suppressing \textit{task-irrelevant} visual content. We therefore ask whether VLMs can effectively ignore such distractors. To test this, we perform a proof-of-concept experiment: we apply targeted blackout manipulations to the input image, masking out regions corresponding to task-irrelevant objects before passing the image to the VLM. For this, we consider two popular VLM models: LLaVA-1.5 \citep{liu2024improved} and Qwen-2.5-VL \citep{bai2025qwen3}, and evaluate them on POPE and CV-Bench. For each question in the benchmark, we extract the referring nouns and use OWLv2~\citep{NEURIPS2023_owlv2}, a strong open-vocabulary object detection foundation model, to localize the relevant objects in the image. All remaining regions, i.e., those not referenced in the question, are blacked out prior to being passed to the VLM. If internal binding mechanisms are robust, removing irrelevant visual content should have little effect on performance, since the model should already selectively attend to the relevant objects. Instead, as shown in \cref{tab:binding_results}, blackout consistently improves performance by 1--3\% on POPE and $\sim$1\% on CV-Bench, confirming that task-irrelevant visual content does interfere with compositional reasoning, even when the dense features of the visual backbone are rich and informative. This suggests that binding failures at the reasoning stage are not just an artifact of weak visual features, but reflect a deeper failure of how the VLM integrates visual information.

\begin{table}

\begin{minipage}{0.54\linewidth}

\caption{
\small{\textbf{Causal mediation analysis.} Replacing background tokens with their blacked-out counterparts consistently improves performance, implicating downstream LLM processing in binding failures. See \cref{subsec:causal_analysis} for discussion.}}
\vspace{0.1cm}
\resizebox{1.0\linewidth}{!}{
\label{tab:causal_patch_analysis}
\setlength{\tabcolsep}{1pt}
\centering
\begin{tabular}{@{} l c c c @{}}
\toprule
\textbf{Model} & \textbf{Condition} & \textbf{POPE} & \textbf{CV-Bench} \\
\midrule
\multirow{2}{*}{LLaVA-1.5}
  & Obj: blackout + BG: orig.          & 81.1 & 65.0 \\
  & Obj: orig. + BG: blackout       & \textbf{81.4} & \textbf{65.2} \\
\addlinespace
\multirow{2}{*}{Qwen-2.5-VL}
  & Obj: blackout + BG: orig.          & 79.1 & 79.8 \\
  & Obj: orig. + BG: blackout       & \textbf{81.5} & \textbf{80.5} \\
\bottomrule
\end{tabular}}
\end{minipage}
\hfill
\begin{minipage}{0.44\linewidth}
\caption{
\small{\textbf{Token Dropping.} Dropping task-irrelevant tokens as an OC mechanism improves POPE but degrades CV-Bench, likely due to disrupted spatial structure. See \cref{subsec:token_drop}.}}
\resizebox{1.0\linewidth}{!}{
\label{tab:token_dropping}
\setlength{\tabcolsep}{2pt}
\centering
\begin{tabular}{@{} l c c c @{}}
\toprule
\textbf{Model} & \textbf{Condition} & \textbf{POPE} & \textbf{CV-Bench} \\
\midrule
\multirow{2}{*}{LLaVA-1.5}
  & Original          & 80.5 & 64.7 \\
  & +Blackout       & \textbf{81.7} & \textbf{65.8} \\
\addlinespace
\multirow{2}{*}{Qwen-2.5-VL}
  & Original          & 78.7 & 79.3 \\
  & +Blackout       & \textbf{81.9} & \textbf{80.2} \\
\bottomrule
\end{tabular}}
\end{minipage}

\end{table}

To qualitatively validate this, we apply Logit lens analysis~\citep{neo2025towards} to trace how the model's prediction evolves across layers pre- and post-blackout. As illustrated in \cref{fig:logit_lens}, for a cluttered indoor scene from POPE, where the model incorrectly predicts `ph(one)' with high confidence ($p=0.98$) on the original image, blacking out task-irrelevant objects shifts the top-decode at layer $31$ to the correct answer ``Remote'' ($p=0.39$), which also translates to correct final LlaVA prediction. This confirms that task-irrelevant visual context \textit{actively suppresses} correct object representations within the LLM, and that removing it enables the model to recover the right answer, \textit{even in} deeply cluttered scenes.

\subsection{Where do these Binding Failures arise: The Visual Encoder or LLM?}
\label{subsec:causal_analysis}

The blackout experiments show that task-irrelevant visual content can interfere with VLM reasoning, but do not reveal where this interference arises. In particular, it could originate from the visual encoder or from the subsequent processing of visual tokens. To localize the source of binding failures within the VLM pipeline, we perform a causal mediation analysis~\citep{visualsymbolic_in_vlm_assouel_iclr2026,meng2022locating} by intervening directly on the patch token representations produced by the visual encoder. Specifically, given an original image and its task-irrelevant blacked-out counterpart, we encode both independently to obtain two sets of patch tokens. We then construct two token-splicing conditions: \textbf{(1)} \textbf{Obj: orig. + BG: blackout}, wherein  \textit{object-relevant} tokens are retained from the original image (``Obj: orig.'') while the background tokens are replaced with those from the blacked-out image (``BG: blackout''), and \textbf{(2)} \textbf{Obj: blackout + BG: orig.}, where background tokens from the original image are preserved and object-relevant tokens are replaced with ones from the blacked-out image. We note that in both these experimental setups, the visual encoder is kept frozen and only the output visual token sequences are manipulated before passing onto the LLM. Therefore, since these interventions operate directly on the frozen visual tokens, any resulting change in performance will explicitly due to how VLM processes and binds the encoded visual information. Hence, it allows us to disentangle whether any binding failures originate in the LLM or not. A detailed schematic of the token-splicing procedure is provided in \cref{fig:token_splicing_schematic} of the supplementary.

Our findings in \cref{tab:causal_patch_analysis} reveal that the LLM is the primary locus of distraction: when background patch tokens are replaced with their blacked-out counterparts (``Obj: orig. + BG: blackout''), the performance gain is consistent across multiple architectures and benchmarks, confirming the hypothesis that eliminating task-irrelevant visual noise from the token sequence passed to LLM leads to improved binding. This highights a rather important point, that the binding problem in VLMs is not simply a visual representation problem, and that \textit{robust} object-centric behavior requires multimodal co-design: addressing binding purely at the level of the visual encoder is\textit{ insufficient} if the LLM remains susceptible to cross-object interference. 

\subsection{Object-Centric Mechanisms as a Lightweight Alternative}
\label{subsec:token_drop}

The above results suggest that explicitly controlling which visual tokens are presented to the LLM can partially mitigate binding failures. To further examine this possibility in a training-free setting, we evaluate token dropping as a lightweight alternative to image-level blackout. Rather than masking image regions prior to visual encoding, token dropping directly removes patch tokens corresponding to task-irrelevant objects before they are passed to the LLM. We find that token dropping yields \textit{larger} gains than image blackout on POPE benchmark (see \cref{tab:token_dropping}), but performs worse on CV-Bench. We hypothesize that this discrepancy arises from the disruption of spatial structure introduced by token dropping, which is more detrimental to CV-Bench due to its greater reliance on spatial relationships for answering questions. These results suggest that fully exploiting token-level object-centric mechanisms require jointly designing the intervention and the underlying VLM, for example by fine-tuning the model to robustly process token-dropped inputs.

Taken together, our findings motivate a shift in how object-centric mechanisms are integrated into multimodal models. Rather than compressing scene content into discrete slots at the potential cost of information loss, object-centric approaches could instead provide explicit mechanisms for selective binding while preserving task-relevant information. This perspective also highlights the importance of evaluating object-centric representations within multimodal architectures, where the interaction between visual representations and language-based reasoning can be directly studied and controlled.
\section{Conclusion}
In this work, we present a systematic study of how object-centric representations support binding in modern foundation VLMs, examining three complementary aspects: scene segregation, object representation, and compositional reasoning. Across these dimensions, we find that current slot-based object-centric methods consistently underperform dense patch-based representations, often introducing information loss and compromising attribute preservation. Our findings further reveal that dense representations exhibit stronger emergent object binding, although they remain susceptible to compositional failures. Taken together, these results highlight important limitations of the increasingly explored paradigm of explicit, discrete slot-based abstraction for binding in foundation models. They motivate a closer examination of how object-level structure can be leveraged without sacrificing the rich visual information preserved in dense representations, opening avenues for approaches that better balance structural reasoning with information preservation.


\section{Limitations}

Our study focuses on slot-based object-centric methods built on pretrained vision backbones, providing a foundation for extending the analysis to alternative paradigms, such as diffusion-based models and temporal reasoning frameworks. Additionally, our evaluations focus on static image benchmarks, with video, embodied, and sequential settings offering important directions for future investigation.

\noindent\makebox[\linewidth]{\rule{\linewidth}{0.1pt}}

\noindent \textbf{Acknowledgment.} This work is supported by a grant from Google and Kotak IISc AI-ML Centre.


\newpage
\bibliographystyle{unsrtnat}
\bibliography{neurips_2026}

\newpage
\appendix


\section*{Appendix}

\section{Extended Related Works}
\label{sec:related_works_supp}

\noindent \textbf{Binding in Deep Learning.}
The binding problem -- associating attributes with the correct entities -- remains a fundamental challenge in deep learning, and is particularly pronounced in modern VLMs. Recent works have highlighted persistent binding failures even in large-scale models: \citet{vlm_limits_campbell_neurips2024} demonstrate that VLMs exhibit capacity limitations rooted in incorrect feature binding, while \citet{visual_structures_help_vlm_izadi_neurips2025} show that structured representations can partially alleviate these issues. Other efforts mitigate hallucinations through object grounding \citep{obj_grounding_vlms_geigle-emnlp2024} or improve VQA performance via object-centric structure \citep{oc_in_vqa_mamaghan_iclr2025}. More recent approaches incorporate explicit structural or symbolic mechanisms, such as CTRL-O \citep{ctrl_o_Didolkar_cvpr2025} and object-centric binding in CLIP \citep{ocbinding_in_clip_assouel_neurips2025}, or study emergent symbolic reasoning in VLMs \citep{visualsymbolic_in_vlm_assouel_iclr2026}. While these works collectively suggest that structured representations are beneficial for binding, they stop short of establishing whether existing object-centric formulations are sufficient to address these failures in practice.

\noindent \textbf{Object-Centric Representation Learning (OCL).}
Object-centric learning aims to decompose scenes into structured object-level representations. Early approaches introduced sequential attention \citep{attend_infer_repeat_eslami_neurips19} or variational frameworks \citep{monet_burgess19, iodine_greff19a_icml19} for unsupervised scene decomposition, later combined with spatial decomposition and generative modeling \citep{space_Lin_iclr2020, genesis_Engelcke_iclr2020}. Slot Attention \citep{slot_attention_locatello_neurips2020} has since become the dominant paradigm, using iterative attention to bind features to a fixed set of slots, with subsequent works extending it to improve compositionality \citep{slate_singh_iclr2022} or model temporal dynamics \citep{scalor_Jiang_iclr2020}. A major limitation of these methods is their reliance on synthetic datasets; DINOSAUR \citep{dinosaur_seitzer_iclr2023} bridges this gap by grounding slot representations in pretrained DINO features, with follow-up works introducing additional inductive biases or self-training strategies \citep{cae_lowe_neurips2023, spot_Kakogeorgiou_cvpr2024}. Despite their conceptual appeal, the effectiveness of these methods in capturing semantically meaningful representations in realistic scenarios remains underexplored. Several works attempt to characterize OCL representations through robustness and invariance \citep{pmlr-dittadi_icml2022}, compositional generalization \citep{kapl2026are}, or proxy evaluations such as segmentation quality and linear probes -- but a direct characterization of the information encoded by object-centric representations in relation to downstream reasoning remains lacking.

\noindent \textbf{OCL in Vision-Language Models.}
Recent efforts have explored incorporating object-centric representations into VLMs. GroupViT \citep{groupvit_Xu_cvpr2022} shows that object-like groupings can emerge from language supervision, while GROUNDHOG \citep{groundhog_Zhang_cvpr2024} and OMG-LLaVA \citep{omg_llava_neurips2024} introduce object-aware reasoning modules in large multimodal systems. SlotDiffusion \citep{slotdiffusion_neurips2023} and Object-Centric Slot Diffusion \citep{oc_slot_diffusion_jiang_neurips2023} integrate slot-based representations with diffusion models for structured generation and editing.

\noindent \textbf{Evaluating OC Representations.}
\citet{oc_in_vqa_mamaghan_iclr2025} evaluate OC representations on real-world VQA tasks by plugging them into LLMs, reporting competitive performance with pretrained vision models such as DINOv2. However, their setup is limited in several key respects: it trains small language models from scratch rather than leveraging pretrained foundation LLMs; both the OC model and the language model are trained on the same dataset used for evaluation, precluding reliable conclusions about generalization; and VQA is formulated as closed-set classification via an MLP, which is misaligned with the open-ended nature of real-world QA.

Concurrently, \citet{singh2026evaluatingobjectcentricmodelsobject} evaluate OC representations in a foundation VLM setting, studying how object-discovery metrics relate to downstream performance. While we share a similar evaluation protocol, our work differs in scope, focus, and conclusions. We ground our analysis in the binding framework of \citet{greff2020binding}, enabling a fine-grained investigation of \textit{which specific representational properties} current OC models acquire or lack, rather than characterizing their general benchmark performance as done by \citet{singh2026evaluatingobjectcentricmodelsobject}. This finer-grained lens leads to different conclusions: while \citet{singh2026evaluatingobjectcentricmodelsobject} find that OC representations remain competitive with dense representations despite using substantially fewer visual tokens, implicitly assuming that increasing the number of slots should narrow this gap, we explicitly test this hypothesis and show that it does not hold -- the performance gap \textit{persists} even as slot count is substantially increased (see \cref{tab:ftdino_eval_slot_count} of the main paper) -- suggesting that the limitation lies in representation quality rather than the number of visual tokens. Going further, we disentangle the roles of scene segregation, representation quality, and compositional reasoning using RSA and causal mediation analysis (\cref{subsec:repr_quality_oc} and \cref{subsec:causal_analysis} of the main paper): while \citet{singh2026evaluatingobjectcentricmodelsobject} report that standard object-discovery metrics correlate poorly with downstream VLM performance and implicitly assume the importance of segmentation quality in their proposed metric without providing direct experimental evidence, we explicitly isolate the effect of segmentation from representation quality and show that scene segregation does in fact causally contribute to downstream reasoning performance. Finally, we extend our analysis beyond OC representations, an aspect not explored by \citet{singh2026evaluatingobjectcentricmodelsobject}, showing that binding failures persist even in dense representations, with the LLM, rather than the visual encoder, emerging as the primary locus of these failures.

\begin{figure}
    \centering
    \includegraphics[width=\textwidth]{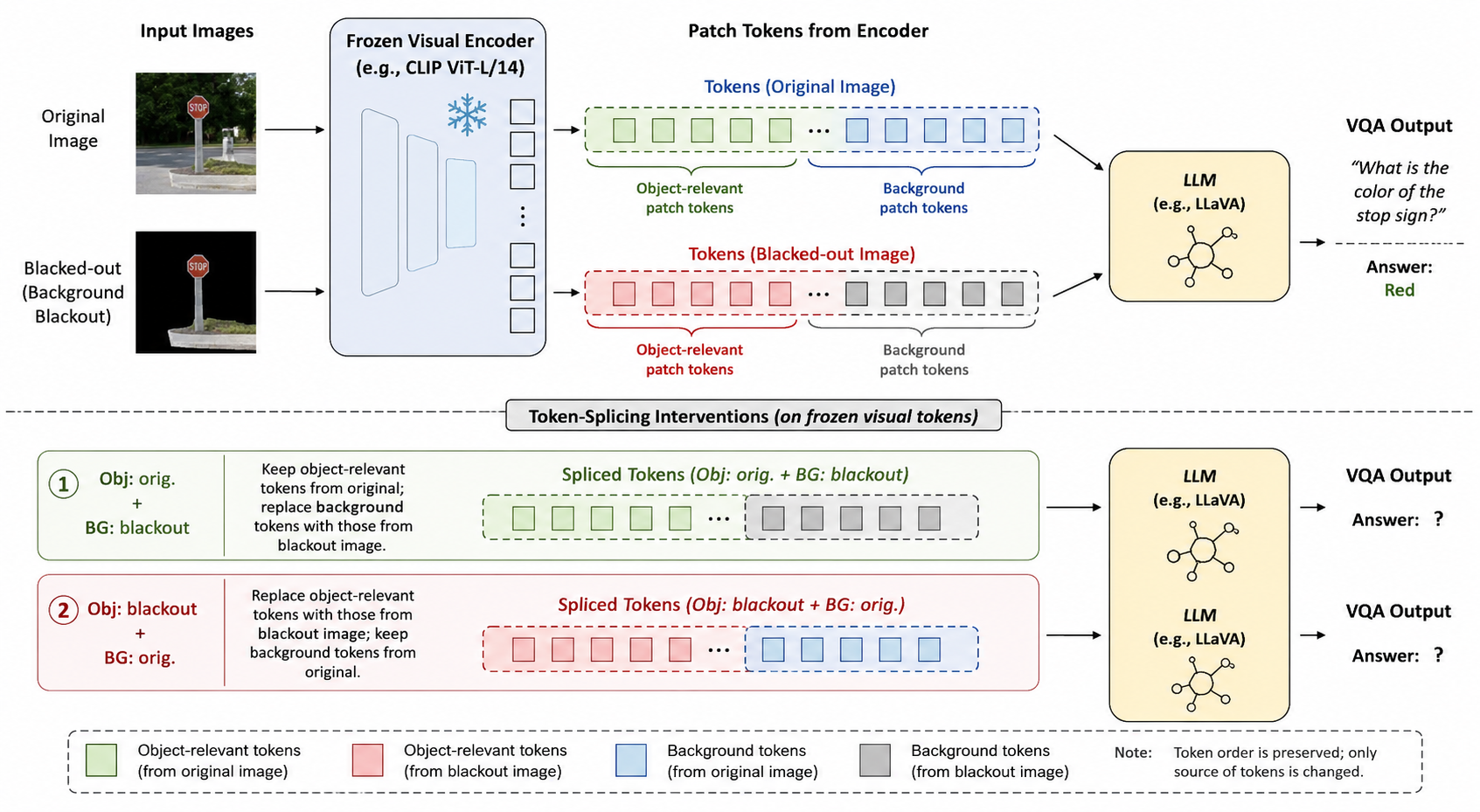}
    \caption{\small{\textbf{Schematic for the token-splicing intervention experiment for localizing binding failures in the VLM (\cref{subsec:causal_analysis} of the main paper).} The original and background-blackout images are independently passed through the frozen visual encoder to obtain their patch-token representations. We then intervene directly on these frozen tokens by swapping either the object-relevant or background tokens between the two images. In \textbf{condition (1)}, object tokens are retained from the original image while background tokens are taken from the blackout image; in \textbf{condition (2)}, the roles are reversed. The resulting spliced token sequences are passed to the LLM for VQA. Since the visual encoder is held fixed and only the encoded token representations are manipulated, changes in model behavior isolate the role of downstream visual-token processing and binding in the LLM.}}

    \label{fig:token_splicing_schematic}
\end{figure}

\section{Training and Evaluation Details}

\noindent \subsection{VLM Instruction Training}
\label{subsec:vlm_instruct_tuning_details_supp}

We follow the standard LLaVA-v$1.5$ instruction-tuning protocol and training configuration. Specifically, the feature alignment stage uses the standard $558$K LAION-CC-SBU image-text pairs, followed by visual instruction finetuning on the standard $665$K-sample LLaVA instruction-tuning mixture (comprising GPT-generated multimodal instructions, VQA datasets, and ShareGPT conversations). In both stages, training uses the standard autoregressive next-token cross-entropy loss, identical to the original LLaVA-v$1.5$ setup.

\noindent \subsection{Vision Encoder Details}

We evaluate a range of vision encoders combined with multiple LLM backbones within the LLaVA framework. All models are trained using the original LLaVA training pipeline (mentioned above) and hyperparameters, without additional tuning, to ensure fair comparison across vision encoders. We consider both pretrained and object-centric learning (OC)-based vision encoders to analyze how different representation paradigms influence downstream multimodal performance.

\begin{itemize}

 \item \textbf{DINOv1 \citep{Caron_2021_ICCV} / DINOv2 \citep{oquab2024dinov}.} 
We use DINOv1 and DINOv2 as pretrained baselines without any architectural or training modifications. These self-supervised vision transformers are trained on large-scale image corpora and are known to produce strong global representations that capture high-level semantic structure. We directly integrate their features into the LLaVA pipeline, treating them as fixed visual backbones. This setup provides a strong reference point for evaluating the effectiveness of object-centric representations.

 \item \textbf{SPOT \citep{spot_Kakogeorgiou_cvpr2024}.} 
SPOT is an unsupervised object-centric learning framework based on slot-attention autoencoding that learns structured object-level representations through attention-guided training and autoregressive reconstruction. The model builds upon pretrained DINOv1 visual features for object decomposition, making it a natural object-centric counterpart to the dense DINOv1 backbone used in our comparisons. In our experiments, we use the pretrained SPOT encoder with an embedding dimension of 256 and standardize the input resolution to $336 \times 336$ for consistent evaluation across all models.

\item \textbf{FT-DINO \citep{didolkar2025on}.} 
FT-DINO is an object-centric representation learning framework that adapts pretrained self-supervised vision encoders for object discovery through task-specific fine-tuning, improving zero-shot transfer across diverse real-world and synthetic datasets. The method builds upon pretrained DINOv2 visual features to produce semantically stronger slot-based object representations. In our setup, FT-DINO generates embeddings of dimension 256 and operates at an input resolution of $336 \times 336$. We directly integrate the pretrained encoder outputs into the LLaVA pipeline without additional fine-tuning.

\item \textbf{DINOSAURv2 \citep{seitzer2023bridging}} 
DINOSAUR is an unsupervised object-centric learning framework designed to discover structured object representations from visual features. In this work, we train a variant, which we refer to as DINOSAURv2, that uses pretrained DINOv2 features as reconstruction targets instead of the original DINOv1 features. We follow the publicly available unofficial implementation\footnote{\texttt{https://github.com/gorkaydemir/DINOSAUR}}
 and adapt it to our setting, configuring the model to produce embeddings of dimension 768 while using a standardized input resolution of 336×336.

\item \textbf{GLASS \citep{glass_Singh_2025_CVPR}.} 
GLASS is an object-centric learning framework that combines slot attention with latent diffusion-based image representations and semantic guidance mechanisms to improve object decomposition in complex real-world scenes. The model builds upon pretrained DINOv2 visual features and learns semantically guided slot embeddings for downstream object discovery and compositional generation tasks. In our experiments, we use the original pretrained checkpoint without retraining. The model outputs embeddings of dimension 1024, and we standardize the input resolution to $336 \times 336$ while keeping all pretrained weights unchanged. 

\end{itemize}

\subsection{Evaluation Benchmarks}

We evaluate all models on a diverse suite of benchmarks spanning multimodal perception, reasoning, grounding, and hallucination robustness. The selected benchmarks collectively measure both low-level visual understanding and high-level multimodal reasoning capabilities.

\begin{itemize}

\item \textbf{CVBench \citep{tong2024cambrian}.}
We evaluate compositional visual understanding using the CVBench benchmark following \textit{VLM Evaluation}\footnote{\texttt{https://github.com/TRI-ML/vlm-evaluation}}. CVBench is designed to assess both 2D and 3D visual reasoning capabilities in multimodal models through a diverse set of perception and spatial reasoning tasks.

For the 2D evaluation, we consider two benchmark subsets: \textit{Bench2D-Count}, consisting of 788 samples focused on object counting, and \textit{Bench2D-Relation}, consisting of 600 samples evaluating relational and spatial reasoning. The final \textit{Accuracy-2D} score is computed as a weighted average across these subsets based on the number of evaluation samples.

For the 3D evaluation, we use two benchmark subsets: \textit{Bench3D-Depth}, containing 600 samples for depth understanding, and \textit{Bench3D-Distance}, containing 600 samples for distance estimation and spatial reasoning. The final \textit{Accuracy-3D} score is computed by averaging the performance across these two subsets.

Following the official evaluation protocol, the overall CVBench score is computed as the average of \textit{Accuracy-2D} and \textit{Accuracy-3D}. We additionally report detailed metrics for all benchmark subsets to provide a comprehensive analysis of multimodal perception and spatial reasoning performance.

\item \textbf{RefCOCO \citep{refcoco_eccv16}.}
We evaluate visual grounding performance on the RefCOCO family of datasets, including RefCOCO val set, RefCOCO+, and RefCOCOg. These benchmarks assess a model’s ability to localize image regions conditioned on natural language referring expressions ranging from simple object descriptions to compositional and context-dependent queries. To ensure computational efficiency while preserving statistical diversity, we evaluate on a curated subset of 3,017 samples distributed across the three benchmarks. This subset maintains broad coverage of referring expression categories, scene compositions, and grounding difficulty while enabling scalable comparison across all evaluated models. To validate the correctness of our evaluation pipeline, we additionally reproduce and match the reported LLaVA baseline results from recent grounding evaluations \citep{jiang2026analyzing}.

\item \textbf{MME \citep{fu2026mme}.}
We evaluate model capabilities using the MME benchmark, which assesses multimodal large language models across a diverse set of perception and cognition tasks. The perception component measures visual understanding through tasks such as object recognition, attribute identification, and OCR, while the cognition component evaluates higher-level reasoning abilities, including commonsense inference, numerical reasoning, and logical understanding. Performance is reported using the overall MME score, which is the sum of the perception and cognition scores.

\item \textbf{POPE \citep{pope_li_etal_emnlp2023}.}
We assess hallucination robustness using the POPE benchmark. It measures object-level perception and hallucination by asking models whether specific objects are present in an image. Following the standard LLaVA evaluation protocol, we evaluate models under three settings: \textit{random}, \textit{popular}, and \textit{adversarial}. In the \textit{random} setting, absent objects are sampled uniformly at random; the \textit{popular} setting selects negative objects from a pool of frequently occurring categories; and the \textit{adversarial} setting focuses on objects that commonly co-occur with those in the image but are intentionally absent, making grounding more challenging. Each setting comprises 3,000 image–question pairs. The final POPE score is computed by averaging performance across all three evaluation settings.

\item \textbf{GQA} \citep{Hudson_2019_CVPR}
We evaluate compositional visual reasoning using the GQA benchmark, which is designed to measure grounded question answering and multi-step reasoning in real-world scenes. Following the standard evaluation protocol, we report overall question answering accuracy on the balanced test split. 
\end{itemize}





\section{Details on Property Prediction task}
\label{suppsec:property_prediction_task}

For all Property Prediction experiments, images are resized to $336 \times 336$. We divide the representations into two groups based on their visual backbone: DINOv1 (DINOv1 and SPOT) and DINOv2 (DINOv2, DINOSAURv2, FT-DINO, and GLASS).

For dense DINO representations, we select the six largest ground-truth COCO object masks in each image and randomly sample two patch representations from each object region, resulting in a representation tensor of shape $(B,12,D)$, where $B$ is the batch size and $D$ is the embedding dimension. For object-centric models, we directly use the original slot representations produced by each method without modifying the number of slots.

We use a linear predictor independently on each slot or patch representation. The predictor outputs class logits and continuous normalized coordinate predictions. We use cross-entropy loss for classification and mean squared error for coordinate regression, with the two losses summed to obtain the training objective. For object-centric representations, predicted slots are matched to ground-truth objects using Hungarian matching to minimize the assignment cost.

We use a batch size of $128$, a learning rate of $10^{-3}$, and train for $10{,}000$ optimization steps. For object-centric methods, we use $20{,}000$ COCO training images for training, $1{,}000$ for validation, and $5{,}000$ COCO validation images for testing. For dense DINOv1 and DINOv2 representations, we use $10{,}000$ training images because they use twice as many visual representations per image. We report results from the checkpoint with the best validation performance.

\section{Mathematical Description of the Patch-Average Algorithm}
\label{suppsec:patch_average}
Given frozen DINOv2 patch features $f_i$, $i=1,\ldots,N$, and object masks $M_j$, $j=1,\ldots,k$, obtained either from a slot encoder or an independent segmentation model, we construct a patch-averaged representation for each object by averaging the features of patches within its mask:
\begin{equation}
z_j = \frac{1}{|M_j|}\sum_{i\in M_j} f_i,
\end{equation}
where $|M_j|$ denotes the number of patches in mask $M_j$. We then use the resulting representations $\{z_j\}_{j=1}^k$ in place of the learned slot representations and pass them to the VLM.

\section{Additional Results on VQA Benchmarks}
\label{suppsec:additional_vqa_results}

\noindent \subsection{MME}

\noindent We present additional details on the MME benchmark \citep{fu2026mme} by separating performance into perception and cognition categories in \cref{tab:supp_mme}. Across all LLM backbones, dense DINO features consistently outperform their OC counterparts on perception-heavy tasks. The gap becomes larger with stronger vision encoders. For instance, under the Qwen2.5-7B backbone, DINOv2 \citep{oquab2024dinov} achieves a perception score of $1399.0$, whereas FT-DINO and DINOSAURv2 obtain only $1064.0$ and $1163.6$, respectively. Similar trends are observed across Phi4 and Vicuna backbones. 

\begin{table*}
\centering
\caption{\textbf{MME perception and cognition breakdown across vision encoders and LLM backbones.} We report Perception (Perc.) and Cognition (Cog.) scores for dense visual features and OCL-based representations across Qwen-2.5, Phi-4, and Vicuna model families. While DINOv2 consistently improves both perception and cognition scores over DINOv1 in the dense setting, the corresponding OC variants fail to preserve these gains.}
\vspace{0.1cm}
\setlength{\tabcolsep}{6pt}
\label{tab:supp_mme}
\resizebox{\linewidth}{!}{
\begin{tabular}{l cc cc cc cc cc}
\toprule
\multirow{2}{*}{\textbf{Encoder}} 
& \multicolumn{2}{c}{\textbf{Qwen2.5-0.5B}} 
& \multicolumn{2}{c}{\textbf{Qwen2.5-3B}} 
& \multicolumn{2}{c}{\textbf{Phi4-3.8B}} 
& \multicolumn{2}{c}{\textbf{Vicuna-7B}} 
& \multicolumn{2}{c}{\textbf{Qwen2.5-7B}} \\
\cmidrule{2-11}
& \textbf{Perc.} & \textbf{Cog.}
& \textbf{Perc.} & \textbf{Cog.}
& \textbf{Perc.} & \textbf{Cog.}
& \textbf{Perc.} & \textbf{Cog.}
& \textbf{Perc.} & \textbf{Cog.} \\
\midrule

\multicolumn{11}{c}{\textbf{Backbone: DINOv1}} \\
\midrule
DinoV1 
& \textbf{916.2} & \textbf{253.9}
& \textbf{1041.9} & 342.5
& \textbf{1118.6} & \textbf{304.3}
& \textbf{1166.9} & 310.7
& \textbf{1184.2} & 305.0 \\

SPOT   
& 915.5 & 241.8
& 1014.3 & \textbf{344.3}
& 1087.6 & 289.6
& 1083.8 & \textbf{311.4}
& 1116.6 & \textbf{325.4} \\

\midrule
\multicolumn{11}{c}{\textbf{Backbone: DINOv1}} \\
\midrule
DinoV2 
& \textbf{1076.5} & \textbf{263.6}
& \textbf{1277.3} & 318.9
& \textbf{1273.1} & 275.7
& \textbf{1314.3} & 346.4
& \textbf{1399.0} & \textbf{358.6} \\

DINOSAURv2 
& 979.9 & 233.6
& 1103.6 & \textbf{334.6}
& 1122.4 & \textbf{298.9}
& 1194.3 & 315.3
& 1163.6 & 341.1 \\

FT-DINO  
& 953.1 & 231.4
& 1048.0 & 323.6
& 1102.0 & 277.5
& 1161.5 & 275.0
& 1064.0 & 305.4 \\

GLASS  
& 982.0 & 244.3
& 1116.5 & 326.4
& 1202.5 & 276.1
& 1290.5 & \textbf{363.2}
& 1159.5 & 322.5 \\

\bottomrule
\end{tabular}}
\end{table*}

In contrast, cognition scores exhibit comparatively smaller differences between dense and object-centric representations. In some cases, OC variants remain competitive on cognition-oriented tasks despite large drops in perception performance. For example, GLASS \citep{glass_Singh_2025_CVPR} achieves a cognition score of $363.2$ with Vicuna-7B \citep{vicuna2023}, close to the dense DINOv2 score of $346.4$, while still underperforming substantially on perception. This suggests that higher-level reasoning capabilities are primarily governed by the language model, whereas accurate perception remains strongly dependent on preserving dense spatial visual features.

Overall, these results indicate that the primary weakness of current OC representations lies in retaining low-level perceptual information rather than abstract reasoning capability. Although object-centric representations provide a compact semantic abstraction of the scene, this compression appears to remove fine-grained visual cues necessary for robust perception and grounding.

\noindent \subsection{POPE}

\noindent We present detailed POPE \citep{pope_li_etal_emnlp2023} results across multiple LLM backbones in \cref{tab:supp_pope_1,tab:supp_pope_2}. Across all settings, dense DINO representations consistently achieve the strongest performance on the Popular, Random, and Adversarial splits, while OCL-based representations exhibit a significant degradation, particularly on adversarial examples. The trend remains stable across both small and large language models, suggesting that the limitation primarily originates from the visual representation rather than the reasoning capacity of the LLM.

The gap becomes pronounced with stronger vision encoders. For example, with the Qwen2.5-7B backbone, dense DINOv2 achieves $88.4/89.1/86.4$ on the Popular/Random/Adversarial splits, whereas DINOSAURv2 achieves only $83.7/85.0/79.2$, and GLASS further drops to $83.9/84.8/79.2$. Similar trends are observed across Phi4 and Vicuna backbones. Notably, the largest performance drop consistently occurs on the adversarial split, which is specifically designed to evaluate hallucination robustness under visually confusing object-presence queries. This suggests that compressing dense visual tokens into a small set of object-centric slots weakens the model's ability to preserve fine-grained evidence required to reject false object attributions.

\begin{table}
\centering
\caption{\textbf{POPE hallucination evaluation across Phi4, Vicuna, and Qwen2.5-7B backbones.} We report accuracy on the Popular, Random, and Adversarial splits of the POPE benchmark for dense DINO features and OC-based representations.}
\vspace{0.1cm}
\setlength{\tabcolsep}{2pt} 
\label{tab:supp_pope_1}
\resizebox{\linewidth}{!}{%
\begin{tabular}{l ccc ccc ccc}
\toprule
\multirow{2}{*}{\textbf{Encoder}} 
& \multicolumn{3}{c}{\textbf{Phi-4 3.8B}} 
& \multicolumn{3}{c}{\textbf{Vicuna-7B}} 
& \multicolumn{3}{c}{\textbf{Qwen2.5-7B}} \\
\cmidrule(lr){2-4} \cmidrule(lr){5-7} \cmidrule(lr){8-10}
& \textbf{Popular} & \textbf{Random} & \textbf{Adversarial} 
& \textbf{Popular} & \textbf{Random} & \textbf{Adversarial} 
& \textbf{Popular} & \textbf{Random} & \textbf{Adversarial} \\
\midrule

\multicolumn{10}{c}{\textbf{Backbone: DINOv1}} \\
\midrule
DinoV1 
& \textbf{81.4} & \textbf{83.4} & \textbf{75.8} 
& \textbf{82.7} & \textbf{83.3} & \textbf{77.9} 
& \textbf{84.6} & \textbf{82.9} & \textbf{78.1} \\

SPOT   
& 81.1 & 83.1 & 75.5
& 79.8 & 82.3 & 74.0
& 81.6 & 80.1 & 74.9 \\

\midrule
\multicolumn{10}{c}{\textbf{Backbone: DINOv2}} \\
\midrule
DinoV2 
& \textbf{87.3} & \textbf{88.3} & \textbf{84.9}
& \textbf{87.2} & \textbf{87.6} & \textbf{85.6} 
& \textbf{88.4} & \textbf{89.1} & \textbf{86.4} \\

DINOSAURv2
& 82.8 & 83.8 & 78.7
& 83.4 & 84.3 & 80.0 
& 83.7 & 85.0 & 79.2 \\

FT-DINO 
& 83.2 & 84.3 & 79.9 
& 83.1 & 84.3 & 79.6
& 84.3 & 83.4 & 79.5 \\

GLASS  
& 81.9 & 83.9 & 78.9 
& 82.3 & 84.0 & 80.4 
& 83.9 & 84.8 & 79.2 \\

\bottomrule
\end{tabular}
}
\end{table}
\begin{table*}
\centering
\caption{\textbf{POPE hallucination evaluation across lightweight Qwen2.5 backbones.} We report results on the Popular, Random, and Adversarial splits using Qwen2.5-0.5B and Qwen2.5-3B language models.}
\vspace{0.1cm}
\setlength{\tabcolsep}{12pt}
\label{tab:supp_pope_2}
\resizebox{\linewidth}{!}{%
\begin{tabular}{l ccc ccc}
\toprule
\multirow{2}{*}{\textbf{Encoder}} 
& \multicolumn{3}{c}{\textbf{Qwen2.5-0.5B}} 
& \multicolumn{3}{c}{\textbf{Qwen2.5-3B}} \\
\cmidrule(lr){2-4} \cmidrule(lr){5-7}
& \textbf{Popular} & \textbf{Random} & \textbf{Adversarial} 
& \textbf{Popular} & \textbf{Random} & \textbf{Adversarial} \\
\midrule

\multicolumn{7}{c}{\textbf{Backbone: DINOv1}} \\
\midrule
DinoV1 
& \textbf{82.6} & \textbf{83.7} & \textbf{75.5} 
& \textbf{83.6} & \textbf{84.4} & \textbf{77.2} \\

SPOT   
& 80.3 & 82.4 & 72.6
& 80.8 & 82.4 & 74.4 \\

\midrule
\multicolumn{7}{c}{\textbf{Backbone: DINOv2}} \\
\midrule
DinoV2 
& \textbf{87.5} & \textbf{88.4} & \textbf{83.6}
& \textbf{87.6} & \textbf{87.9} & \textbf{85.7} \\

DINOSAURv2
& 83.2 & 84.3 & 77.7 
& 83.9 & 84.3 & 79.6 \\

FT-DINO  
& 83.7 & 84.7 & 77.0 
& 84.8 & 84.6 & 79.2 \\

GLASS  
& 81.6 & 82.5 & 74.6 
& 83.1 & 84.4 & 78.6 \\

\bottomrule
\end{tabular}
}
\end{table*}

Interestingly, upgrading the language model alone does not mitigate this gap. Even with stronger LLMs such as Qwen2.5-7B and Vicuna-7B, OC representations remain substantially below their dense counterparts. In several cases, DINOv1 \citep{Caron_2021_ICCV} dense features outperform DINOv2-based OCL methods despite using a weaker vision encoder. For instance, under Phi4-3.8B \citep{abdin2024phi4technicalreport}, DINOv1 achieves $75.8$ on the adversarial split, while GLASS obtains only $78.9$ despite being built on top of DINOv2. This indicates that improvements in backbone quality are not effectively transferred through the current object-centric bottleneck.

Overall, the POPE results reinforce the conclusion that current OCL methods discard visual details that are essential for reliable object verification and hallucination suppression. While object-centric representations provide a compact semantic abstraction, they appear insufficient for preserving the dense spatial evidence necessary for robust visual grounding under challenging conditions.

\begin{table*}
\centering
\caption{\textbf{CVBench evaluation across Phi4, Vicuna, and Qwen2.5-7B backbones.} We report performance on 2D Counting (2D Count), 2D relational reasoning (2D Rel.), 3D depth estimation (3D Depth), and 3D distance understanding (3D Dist.) for dense DINO features and OCL-based representations.}
\vspace{0.1cm}
\setlength{\tabcolsep}{1.5pt} 
\label{tab:supp_cvbench_1}
\resizebox{\linewidth}{!}{%
\begin{tabular}{l cccc cccc cccc}
\toprule
\multirow{2}{*}{\textbf{Encoder}} & \multicolumn{4}{c}{\textbf{Phi-4 3.8B}} & \multicolumn{4}{c}{\textbf{Vicuna-7B}} & \multicolumn{4}{c}{\textbf{Qwen2.5-7B}} \\
\cmidrule(lr){2-5} \cmidrule(lr){6-9} \cmidrule(lr){10-13}
& \textbf{2D Count} & \textbf{2D Rel.} & \textbf{3D Depth} & \textbf{3D Dist.} & \textbf{2D Count} & \textbf{2D Rel.} & \textbf{3D Depth} & \textbf{3D Dist.} & \textbf{2D Count} & \textbf{2D Rel.} & \textbf{3D Depth} & \textbf{3D Dist.} \\
\midrule
\multicolumn{13}{c}{\textbf{Backbone: DINOv1}} \\
\midrule
DinoV1 & \textbf{42.1} & \textbf{66.6} & \textbf{69.6} & \textbf{57.6} & \textbf{48.6} & \textbf{63.4} & 57.8 & 49.2 & \textbf{42.3} & \textbf{65.7} & \textbf{73.2} & \textbf{65.8} \\
SPOT   & 40.2 & 58.6 & 65.5 & 54.8 & 46.1 & 53.8 & \textbf{68.3} & \textbf{53.5} & 38.0 & 64.8 & 70.5 & 64.3 \\
\midrule
\multicolumn{13}{c}{\textbf{Backbone: DINOv2}} \\
\midrule
DinoV2 & \textbf{56.1} & \textbf{67.1} & \textbf{72.8} & 56.7 & \textbf{54.6} & \textbf{65.5} & 68.8 & 48.8 & \textbf{50.5} & \textbf{70.9} & 73.7 & 62.0 \\
DINOSAURv2 & 43.8 & 61.2 & 69.8 & \textbf{58.2} & 54.4 & 60.2 & 62.5 & 49.3 & 40.7 & 66.2 & 72.5 & 64.8 \\
FT-DINO  & 43.4 & 62.9 & 72.0 & 55.8 & 52.5 & 57.2 & 62.5 & 50.2 & 41.8 & 67.7 & \textbf{74.2} & 62.0 \\
GLASS  & 46.4 & 61.8 & 68.0 & 54.3 & 48.6 & 58.9 & \textbf{70.0} & \textbf{53.0} & 40.0 & 64.3 & 74.0 & \textbf{65.5} \\
\bottomrule
\end{tabular}
}
\end{table*}

\begin{table*}
\centering
\caption{\textbf{CVBench evaluation across lightweight Qwen2.5 backbones.} We report results on 2D and 3D reasoning categories using Qwen2.5-0.5B and Qwen2.5-3B language models.}
\vspace{0.1cm}
\setlength{\tabcolsep}{4pt}
\label{tab:supp_cvbench_2}
\resizebox{\linewidth}{!}{%
\begin{tabular}{l cccc cccc}
\toprule
\multirow{2}{*}{\textbf{Encoder}} 
& \multicolumn{4}{c}{\textbf{Qwen2.5-0.5B}} 
& \multicolumn{4}{c}{\textbf{Qwen2.5-3B}} \\
\cmidrule(lr){2-5} \cmidrule(lr){6-9}
& \textbf{2D Count} & \textbf{2D Rel.} & \textbf{3D Depth} & \textbf{3D Dist.} 
& \textbf{2D Count} & \textbf{2D Rel.} & \textbf{3D Depth} & \textbf{3D Dist.} \\
\midrule

\multicolumn{9}{c}{\textbf{Backbone: DINOv1}} \\
\midrule
DinoV1 
& 17.1 & 49.4 & \textbf{53.8} & \textbf{49.5} 
& \textbf{39.3} & 54.0 & \textbf{57.3} & \textbf{56.8} \\

SPOT   
& \textbf{17.3} & \textbf{49.5} & 50.8 & 48.2 
& 34.4 & \textbf{55.2} & 53.2 & 54.5 \\

\midrule
\multicolumn{9}{c}{\textbf{Backbone: DINOv2}} \\
\midrule
DinoV2 
& \textbf{26.8} & \textbf{52.8} & 51.7 & 48.7
& \textbf{53.7} & \textbf{56.0} & 61.8 & 55.3 \\

DINOSAURv2 
& 22.3 & 51.8 & 54.0 & 54.3
& 37.3 & 54.9 & \textbf{64.3} & 57.2 \\

FT-DINO  
& 17.5 & 50.9 & 53.8 & 47.8
& 38.7 & 55.8 & 62.3 & \textbf{58.8} \\

GLASS  
& 23.9 & 50.5 & \textbf{59.5} & \textbf{55.5}
& 42.0 & 54.3 & 64.2 & 57.8 \\

\bottomrule
\end{tabular}
}
\end{table*}

\noindent \subsection{CV Bench}

\noindent We evaluate dense and object-centric visual representations on CVBench \citep{tong2024cambrian} in \cref{tab:supp_cvbench_1,tab:supp_cvbench_2}, which measures fine-grained 2D and 3D visual reasoning capabilities across counting, relational understanding, and depth perception tasks. Across nearly all LLM backbones and evaluation categories, dense DINO representations consistently outperform their OC counterparts. The performance gap is especially pronounced on tasks requiring accurate spatial localization and preservation of detailed scene structure, with the only exception of 3D reasoning, where OC methods outperform dense features.

The largest degradation is typically observed on 2D counting and relational reasoning tasks. For example, under the Phi4-3.8B backbone, DINOv2 achieves $56.1$ on 2D counting, whereas DINOSAURv2 and FT-DINO obtain only $43.8$ and $43.4$, respectively. Similar trends hold across Qwen2.5 and Vicuna backbones. 
The degradation is also evident on relational reasoning tasks involving spatial configurations. Under Vicuna-7B, DINOv2 achieves $65.5$ on 2D relation understanding, while GLASS drops to $58.9$. These results suggest that current OC representations struggle to retain fine-grained geometric information necessary for relational reasoning between objects. Since slot-based representations prioritize semantic abstraction, spatial precision may be partially lost during the object grouping process.

Interestingly, the gap is comparatively smaller on certain 3D reasoning categories. In some settings, OC methods remain competitive on 3D distance tasks despite underperforming substantially on 2D counting and localization. For instance, under Qwen2.5-3B, FT-DINO achieves $58.8$ on 3D distance compared to $55.3$ for dense DINOv2. This indicates that some coarse scene-level structural information may still be preserved within object-centric representations, particularly for higher-level relational reasoning tasks that depend less on exact spatial detail.

The overall trend remains consistent across both small and large language models: improvements in the underlying vision backbone are not effectively transferred through current OC pipelines. In several cases, dense DINOv1 features remain competitive with or outperform DINOv2-based OC methods despite using a weaker pretrained encoder. 

\noindent \subsection{GQA}

We report GQA \citep{Hudson_2019_CVPR} across multiple Qwen2.5 language model scales in \cref{tab:supp_gqa}. Similar to the trends observed on MME, POPE, and CVBench, dense DINO representations consistently outperform object-centric representations across all settings. The gap becomes larger with stronger vision backbones and larger language models, indicating that current OC approaches are unable to effectively preserve the gains provided by improved pretrained visual features.

For instance, under the Qwen2.5-7B backbone, dense DINOv2 achieves an accuracy of $61.9$, whereas DINOSAURv2, FT-DINO, and GLASS obtain only $54.5$, $53.7$, and $54.4$, respectively. Similar trends are observed for the Qwen2.5-3B backbone, where DINOv2 reaches $60.4$, substantially outperforming all OC variants. Notably, several DINOv2-based OC methods perform only marginally better than, or remain comparable to, dense DINOv1 features despite relying on a significantly stronger pretrained encoder. 

Unlike counting-focused tasks such as CVBench, GQA additionally requires understanding object attributes, relationships, and multi-step compositional reasoning. The consistent degradation across all OC variants indicates that compressing dense patch features into a limited set of object-centric slots may remove contextual cues necessary for fine-grained relational understanding. Although slot representations capture coarse semantic object structure, they appear less effective at preserving the detailed interactions and spatial dependencies required for robust question answering.

Interestingly, the relative performance gap remains stable across different LLM scales. Increasing the language model capacity from Qwen2.5-0.5B to Qwen2.5-7B improves overall performance for both dense and OC representations.

\section{Additional VQA results on MLLM-based Vision backbones}
\label{suppsec:clip_vqa_results}

\begin{table}
\begin{minipage}{0.35\linewidth}

\caption{\small{\textbf{GQA evaluation of OC vs Dense representations} across Qwen$2.5$ language model backbones.}}
\vspace{0.1cm}
\setlength{\tabcolsep}{9pt}
\label{tab:supp_gqa}
\begin{adjustbox}{max width=\linewidth, center}
\begin{tabular}{l c c c}
\toprule
\textbf{Encoder}
& $\mathbf{0.5B}$ 
& $\mathbf{3B}$
& $\mathbf{7B}$ \\
\midrule

\multicolumn{4}{c}{\textbf{Backbone: DINOv1}} \\
\midrule
DINOv1
& \textbf{49.4} & \textbf{52.9} & 51.5 \\

SPOT  
& 46.5 & 51.1 & \textbf{52.5} \\

\midrule
\multicolumn{4}{c}{\textbf{Backbone: DINOv2}} \\
\midrule
DINOv2
& \textbf{54.0} & \textbf{60.4} & \textbf{61.9} \\

DINOSAURv2 
& 49.5 & 53.1 & 54.5 \\

FT-DINO 
& 49.4 & 52.7 & 53.7 \\

GLASS
& 49.0 & 51.1 & 54.4 \\

\bottomrule
\end{tabular}
\end{adjustbox}
\end{minipage}
\hfill
\begin{minipage}{0.63\linewidth}
\caption{
\small{\textbf{VQA Performance comparison of OC methods against their dense counterparts on MLLM-based vision backbones.} Dense  features outperform OC across all evaluated VQA benchmarks.}}
\vspace{0.1cm}
\label{tab:vision_encoder_comparison}
\setlength{\tabcolsep}{2pt}
\centering
\begin{adjustbox}{max width=\linewidth, center}
\begin{tabular}{@{} l c c c c c @{}}
\toprule
\textbf{Vision Encoder} & \textbf{MME} & \textbf{POPE} & \textbf{CV-Bench} &
\textbf{RefCOCO} & \textbf{GQA} \\
\midrule
\multicolumn{6}{c}{\textbf{Qwen2.5-VL-0.5B}} \\
\midrule
Dense (CLIP Orig.)
& \textbf{1382.8}
& \textbf{82.2}
& \textbf{44.3}
& \textbf{12.0}/5.5/\textbf{10.5}
& \textbf{52.5} \\

DINOSAUR (CLIP+OC)
& 1268.9
& 79.0
& 43.1
& 10.5/\textbf{6.1}/9.8
& 48.1 \\

\midrule
\multicolumn{6}{c}{\textbf{Qwen2.5-VL-3B}} \\
\midrule
Dense (CLIP Orig.)
& \textbf{1513.8}
& \textbf{83.7}
& \textbf{53.8}
& \textbf{25.9/17.7/20.4}
& \textbf{55.8} \\

DINOSAUR (CLIP+OC)
& 1443.8
& 80.3
& 51.5
& 21/14.5/18
& 51.9 \\

\bottomrule
\end{tabular}
\end{adjustbox}
\end{minipage}

\end{table}

Most existing object-centric (OC) methods are developed using SSL-based vision backbones which are trained in a language-free manner. To assess the extensibility of our findings to language-aware vision backbones, we additionally evaluate OC representations with CLIP \citep{pmlr-v139-radford21a}, a widely used language-aware vision backbone in modern MLLMs. We train DINOSAUR \citep{seitzer2023bridging} with a frozen CLIP ViT-B/$16$ backbone and use the resulting OC representations to instruction-tune Qwen$2.5$-VL $0.5$B and $3$B models. As a direct baseline, we instruction-tune the same models using dense representations from the same CLIP backbone, following the identical instruction-tuning protocol mentioned in \cref{sec:main_eval} of the main paper.

\cref{tab:vision_encoder_comparison} reports the VQA results across MME, POPE, CV-Bench, RefCOCO, and GQA. Consistent with our SSL-backbone experiments, dense representations consistently outperform OC representations across both model sizes and all evaluated benchmarks. This suggests that our findings extend beyond SSL backbones to CLIP-based MLLM settings. We also report the OC results against the simple patch-averaging baseline in \cref{tab:clip_vqa_patchaverage_05b}. As shown, patch-averaged CLIP representations also outperform the OC representations across all evaluated VQA benchmarks. While patch averaging incurs some expected performance degradation compared to the original dense representation, it remains consistently stronger than the OC representation. 

\begin{table}[t]
\begin{minipage}{0.62\linewidth}
\centering
\caption{
\small{\textbf{VQA performance of OC methods with their patch-averaged counterparts on MLLM-based vision backbones (CLIP).} Results on Qwen$2.5$-VL-$0.5$B. Our findings from \cref{subsec:repr_quality_oc} of the main paper extend to MLLM-based backbones as well.}}
\vspace{0.1cm}
\setlength{\tabcolsep}{3pt}
\label{tab:clip_vqa_patchaverage_05b}
\centering
\begin{adjustbox}{max width=\linewidth, center}
\begin{tabular}{l c c c c c}
\toprule
\textbf{Vision Encoder} & \textbf{MME} & \textbf{POPE} & \textbf{CV-Bench}
& \textbf{RefCOCO} & \textbf{GQA} \\
\midrule

Patch-average (CLIP)
& 1304.0 & 80.6 & 44.0
& 9.4/4.7/8.1 & 48.1 \\

DINOSAUR (CLIP+OC)
& 1268.9 & 79.0 & 43.1
& 10.5/6.1/9.8 & 48.1 \\

\bottomrule
\end{tabular}
\end{adjustbox}
\end{minipage}
\hfill
\begin{minipage}{0.36\linewidth}
\centering
\caption{
\small{\textbf{OC segmentation quality} on unsupervised object discovery benchmarks (FG-ARI, mBO) (\cref{subsec:good_oc_method} of main paper). Results taken from \cite{didolkar2025on}.}}
\vspace{0.1cm}
\setlength{\tabcolsep}{8pt}
\label{tab:supp_ar_mbo}
\centering
\begin{adjustbox}{max width=\linewidth, center}
\begin{tabular}{l c c}
\toprule
\textbf{Model} & \textbf{FG-ARI} & \textbf{mBO} \\
\midrule
DINOSAURv2
& 40.5
& 27.7 \\

FT-DINO
& \textbf{48.8}
& \textbf{36.3} \\

\bottomrule
\end{tabular}
\end{adjustbox}
\end{minipage}

\end{table}




\end{document}